\documentclass[pdflatex,sn-mathphys-ay,iicol]{sn-jnl}% Math and Physical Sciences Numbered Reference Style
\usepackage{graphicx}%
\usepackage{multirow}%
\usepackage{amsmath,amssymb,amsfonts}%
\usepackage{amsthm}%
\usepackage{mathrsfs}%
\usepackage[title]{appendix}%
\usepackage{xcolor}%
\usepackage{textcomp}%
\usepackage{manyfoot}%
\usepackage{booktabs}%
\usepackage{algorithm}%
\usepackage{algorithmicx}%
\usepackage{algpseudocode}%
\usepackage{listings}%
\usepackage{amsmath,amsfonts}
\usepackage{algorithm}
\usepackage{array}
\usepackage[caption=false,font=normalsize,labelfont=sf,textfont=sf]{subfig}
\usepackage{textcomp}
\usepackage{stfloats}
\usepackage{url}
\usepackage{verbatim}
\usepackage{graphicx}
\usepackage{algorithm}
\usepackage{hyperref}
\usepackage{algorithm}
\usepackage{algpseudocode}
\usepackage{amsmath}
\usepackage{amssymb}

\usepackage{subcaption}
\usepackage{tabularx}
\usepackage{booktabs}
\usepackage{tabularx}
\usepackage{multirow}
\usepackage{multirow}
\usepackage{float}
\usepackage{tcolorbox}
\tcbuselibrary{listings,skins,breakable}

\definecolor{llmthink}{RGB}{0,70,180} % blue
\definecolor{llmresp}{RGB}{180,0,0}   % red
\definecolor{sgcolor}{RGB}{0,120,0}   % green
\definecolor{lightgray}{gray}{0.95}
\definecolor{user}{RGB}{0,0,0}

\theoremstyle{thmstyleone}%
\theoremstyle{thmstyletwo}%

\theoremstyle{thmstylethree}%

\begin{document}

\title[sgcot]{Scene Graph-Based Ambiguity Resolution in Robotic Task Planning}


\author*[1]{\fnm{Akshat} \sur{Rana}}\email{akshat.rana.ug22@nsut.ac.in }

\author[1]{\fnm{Peeyush} \sur{Agarwal}}\email{peeyush.agarwal.ug22@nsut.ac.in}

\author[1]{\fnm{K.P.S.} \sur{Rana}}\email{kpsrana@nsut.ac.in}
\author[1]{\fnm{Amarjit} \sur{Malhotra}}\email{amarjit.malhotra@nsut.ac.in}

\affil*[1]{\orgname{Netaji Subhas University of Technology}, \orgaddress{\city{Dwarka}, \postcode{110078}, \state{Delhi}, \country{India}}}

% \affil[2]{\orgdiv{Department}, \orgname{Organization}, \orgaddress{\street{Street}, \city{City}, \postcode{10587}, \state{State}, \country{Country}}}

% \affil[3]{\orgdiv{Department}, \orgname{Organization}, \orgaddress{\street{Street}, \city{City}, \postcode{610101}, \state{State}, \country{Country}}}


\abstract{Ambiguity poses a major challenge to large language models (LLMs) used as robotic planners. In this paper, we present Scene Graph-Chain-of-Thought (SG-CoT), a two-stage framework where LLMs iteratively query a scene graph representation of the environment to detect and clarify ambiguities. First, a structured scene graph representation of the environment is constructed from input observations, capturing objects, their attributes, and relationships with other objects. Second, the LLM is equipped with retrieval functions to query portions of the scene graph that are relevant to the provided instruction. This grounds the reasoning process of the LLM in the observation, increasing the reliability of robotic planners under ambiguous situations. SG-CoT also allows the LLM to identify the source of ambiguity and pose a relevant disambiguation question to the user or another robot. Extensive experimentation on simulations demonstrates that SG-CoT consistently outperforms prior methods, with minimum improvements of 4\% in ambiguity detection and 4\% and 15\% in success rates in single-agent and multi-agent environments respectively.}




\maketitle

\section{Introduction}\label{sec1}

Large Language Models (LLMs) have transformed the landscape of artificial intelligence, providing impressive generalization across domains. Naturally, this progress has also been extended to robotics, where LLMs have been used to achieve reliable human-robot interaction for tasks such as manipulation \citep{rt22023arxiv}, navigation \citep{zhou2023navgpt, chen2024mapgpt} and planning \citep{saycan2022arxiv,huang2022inner}. However, the robots may face ambiguous situations arising from either the state of the environment or underspecification by the user itself. For example, an instruction such as \textit{``Pick up the cup"} becomes ambiguous if there are multiple cups or no cups in the environment, while an instruction like \textit{``Get me something to eat"} is ambiguous because it may lead to multiple valid interpretations. In such cases, a reliable agent must learn to ask for clarification from the user or another robot performing the same task as it may otherwise lead the robot into taking suboptimal or unsafe actions \citep{knowno,ivanova-etal-2025-ambik}.

Since LLMs are inherently overconfident and are not explicitly trained to handle ambiguity \citep{kim-etal-2024-aligning}, it becomes a fundamental challenge for AI-Enabled robotics. Previous approaches \citep{knowno,CLARA} estimate the uncertainty levels in the responses of LLMs to detect ambiguity. However, their evaluation typically focuses on ambiguities in the instruction instead of ambiguities induced by the state of the environment. They also rely on vision-language models (VLMs) like ViLD  \citep{gu2021open} or BLIP  \citep{li2022blip} for perception. Representations generated by these VLMs are often just a list of objects or a high-level description, making it hard for them to disambiguate complex instructions referring to a specific attribute (e.g., \textit{``Get me the object with green buttons"}) or a relation among objects (e.g., \textit{``Get me the object to the left of the apple"}). 

To bridge the above mentioned gap, a Scene Graph-Chain-of-Thought (SG-CoT) is proposed in this paper. SG-CoT is a robotic planning framework that integrates scene graphs with the reasoning capabilities of LLMs to enhance robustness against ambiguity. The framework works in two stages. First, a structured scene graph of the environment is built using a VLM capturing semantic attributes and relationships among objects. This representation moves beyond high-level descriptions and reveals environmental nuances, which are crucial for detecting ambiguities in various scenarios. Secondly, the LLM is paired with a tool-use capability. The LLM receives the high-level descriptions of its observations and the instruction as input. It then begins its thought process and calls a valid API function to access nodes and edges of the scene graph, required to complete the instruction, grounding its thoughts into visual context. Upon retrieval of all relevant information, the LLM makes a decision on whether multiple valid interpretations of the user instruction exist or not. In this way, ambiguities arising from both the environment and underspecification by the user can be resolved. In contrast to purely text-based planners, SG-CoT provides integration of scene graphs with iterative reasoning, allowing the LLM  to pinpoint the reason for the ambiguity and hence generate clarification questions targeted towards resolving the ambiguity.

Two experiments were conducted to validate the proposed method. First, evaluation was performed on simulated tabletop environments presented by SayCan  \citep{saycan2022arxiv}. Second, the methodology was tested on multi-robot tasks derived from the LEMMA benchmark  \citep{Lemma}. To these environments, ambiguity across three different categories was introduced by applying perturbations to both the environment states and the instructions, removing contextual information from the robot’s view. Based on the presented results, it was observed that SG-CoT demonstrated superior performance compared to previous approaches in both executing tasks and asking the correct clarification questions whenever required.

\section{Related work}\label{sec2}

The literature survey has been organized into three components, namely task planning with LLMs, ambiguity in LLMs and scene graphs, each of which aligns with a central element of the proposed method. The survey is concluded with the summary of our main contributions.

\textit{Task planning with LLMs:} Robotic task planning \citep{TaskandMot1} involves translating natural language commands into executable action sequences based on input and past observations. There have been several previous approaches \citep{huang2022inner,CLARA,saycan2022arxiv,ProgPrompt,cap,song2023llmplanner,huang2022language} that use LLMs for this purpose. To increase reliability of LLMs as planners, ProgPrompt \citep{ProgPrompt} used a programmatic prompt structure, LLM-Planner \citep{song2023llmplanner} used a framework for dynamic grounded re-planning and ReAct \citep{yao2022react} interleaved reasoning traces with actions. To further enhance reliability, LLMs employed feedback loops from the environment \citep{huang2022inner,Text2R,FeedbackGrounding} to detect failures and iteratively refine their plans. However,  these approaches assume the environment contains sufficient context to infer a definitive action and will execute an action regardless of the state of the environment. As mentioned in KnowNo \citep{knowno}, such behavior can cause actions with clear safety hazards, like placing a metal bowl in the microwave. 

\textit{Ambiguity resolution in LLMs:} Several prior works have proposed methodologies to tackle the inability of LLMs to handle ambiguity \citep{CLAM,ECLAIR,intent-sim,ambLLM}. For robotics in particular, most previous research focuses on underspecification by the users. KnowNo \citep{knowno} used conformal prediction to generate a statistically guaranteed set of plausible actions, asking for help when this set contains more than one option. CLARA \citep{CLARA}, which classifies user commands as clear, ambiguous, or infeasible through uncertainty estimated via stochastic context sampling, is used as the basis for the problem formulation, with focus restricted exclusively to ambiguity. Furthermore, the problem is expanded beyond simple user underspecification to address a broader class of ambiguities: those arising from the environment itself, the inherent uncertainty of partially observable environments, and those occurring in more complex instructions.

\textit{Scene graphs:} A scene graph is a semantic representation of a scene in which the environment is modeled as a set of objects, their attributes, and the relationships between them \citep{SGsurvey}. They are applied in computer vision for image retrieval \citep{Johnson_2015_CVPR}, image captioning \citep{SG_image_caption,Nguyen_2021_ICCV}, visual question answering \citep{sceneGraphVQA} and image generation \citep{sceneGraphGen}. Various works have also explored their application in robotics \citep{SGnoLLM,SGnoLLM2,SGnoLLM3} and more recent approaches have integrated them with LLMs for generation and understanding  \citep{bird,sgbot,roboexp,verigraph,rana2023sayplan,yin2024sgnav}. The proposed framework is built upon graph-grounded LLM planners like SayPlan \citep{rana2023sayplan}, that pairs an LLM with an API allowing it to expand and collapse a hierarchical 3D scene graph to dynamically manage its own context. However, unlike SG-CoT, SayPlan does not ask for clarification, making it less reliable in ambiguous situations. In addition, it assumes a pre-constructed scene graph and does not support incorporation of open-vocabulary representations of the graph structure.

Based on the above mentioned three components, this paper makes three primary contributions. First, it proposes SG-CoT, a framework that integrates scene graph construction with iterative chain-of-thought reasoning by an LLM to enhance robotic task planning under ambiguous situations. Secondly, ambiguous scenarios are formally classified into three distinct types, including environment-induced ambiguities, which are largely unexplored by prior approaches. Third, the correctness of clarification questions generated by the LLM is measured to determine whether they successfully pinpoint the source of ambiguity.

\section{Proposed Methodology}\label{sec3}

\begin{figure*}[t]
\centering
\makebox[1.1\textwidth][c]{\includegraphics[width=1.2\textwidth]{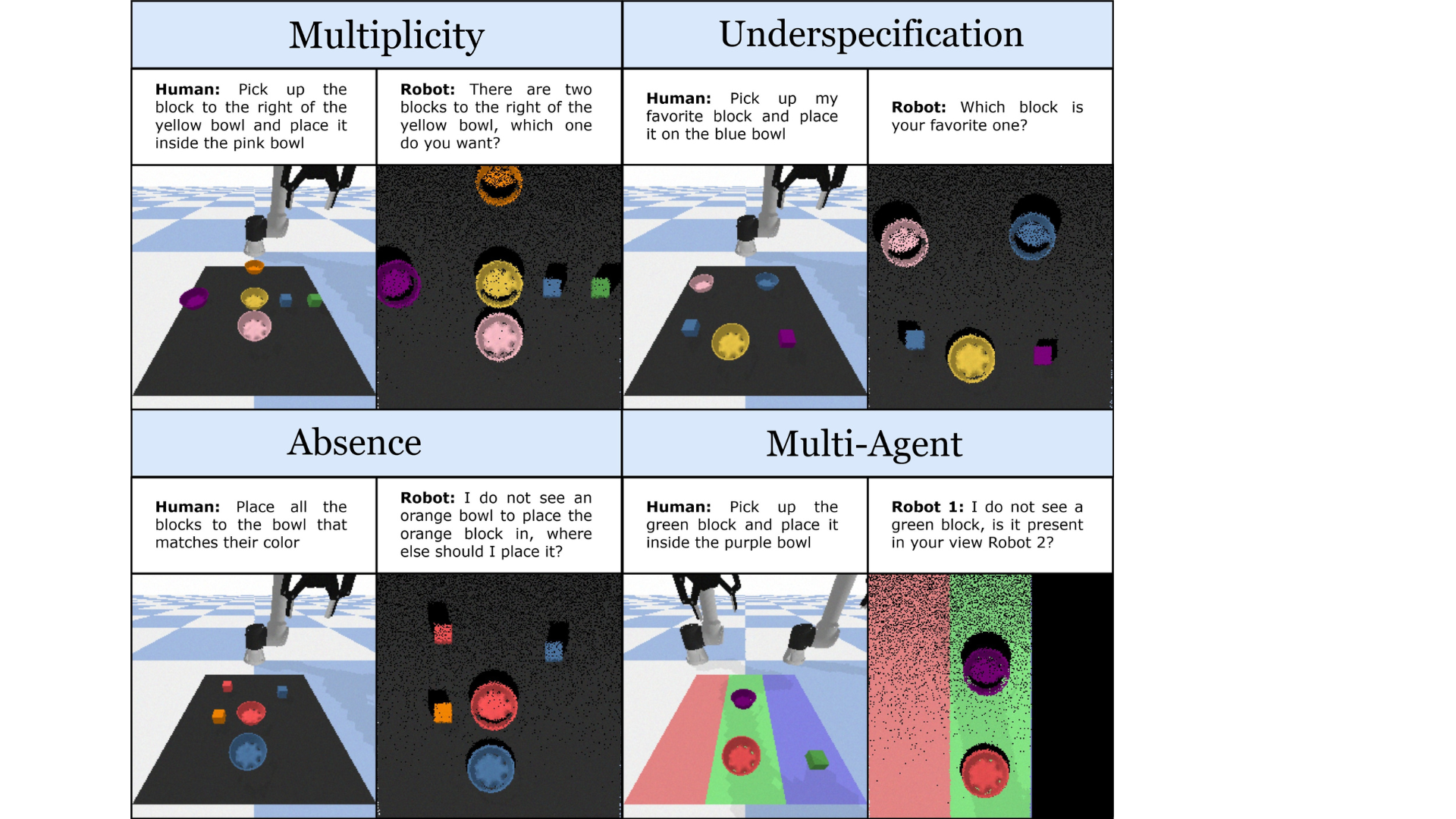}}
\caption{The figure shows possible configurations of the defined ambiguity types. For each category, the image of the left is the global perspective and the human instruction, and the one on the right is the egocentric view of the robot and its clarification question.}
\label{fig:amb}
\end{figure*}

In this section, the proposed framework, referred to as SG-CoT, is introduced. The problem statement is presented first, followed by an introduction of scene graph generation and representation, and finally, the reasoning process that utilizes the scene graph for disambiguation and interaction.

\subsection{Problem Statement}
The problem setup involves a single or multi-agent robotic system integrated with an LLM as a high-level planner for each individual robotic agent $R_i$. Given a natural language instruction $I$ and an image observation $O$ from the environment, the objective of the system is to generate a sequence of high-level actions $a_{i} \in \mathcal{A}$, where $\mathcal{A}$ denotes the predefined set of executable actions available to the robot, to accomplish the defined task. To evaluate the planner’s robustness in handling ambiguity, controlled perturbations were introduced in both the instruction and the environment, as shown in Fig.~\ref{fig:amb}. Instruction-level ambiguity, referred to as underspecification, involves omission of explicit referential grounding in the instruction.  For example, replacing a grounded reference
(e.g., \textit{``place the red block over the green bowl"}) with an ungrounded
term (e.g., \textit{``place something over the green bowl"}). Environment-level ambiguity relates $I$ to the set of objects $X$ present in $O$. Let $A(I)$ be the set of attributes describing a target object referenced in $I$, $A(o)$ be the set of attributes of an object o and let $X_{match} = \{ o \in X_{scene} \mid A(o) \supseteq A(I) \}$ be the set of objects in the scene satisfying these properties. Absence is defined as the condition $X_{match} = \emptyset$, and multiplicity as the condition $|X_{match}| > 1$. Multi-agent systems are also subject to observation-level ambiguities occurring from the partially observed nature of the environment. Let each robot $R_i$ be able to observe only a subset $O_{R_i} \subseteq O$ of the environment, with $X^{R_i} \subseteq X$ being the set of objects visible to $R_i$. The condition $|X_{match}^{R_i}| = 0$ and $|X_{match}| \geq 1$ leads to this type of ambiguity.

\subsection{Scene Graph Generation}
A scene graph provides a structured representation of the environment, derived from the observation $O$. Formally, the scene graph is defined as $G = (V, E)$, where $V = \{v_1, v_2, \dots, v_n\}$ represents the set of nodes, each corresponding to an object $o \in X_{scene}$ detected in the environment. Each node $v_i$ is associated with a set of attributes $A(v_i) = \{a_1, a_2, \dots, a_k\}$ that describe its observable properties, such as color, size, or higher-level states like \textit{is\_broken} or \textit{is\_open}. The edge set $E$ captures pairwise relations between objects, where each edge is represented as a tuple $(v_i, r_{ij}, v_j) \in E$, where $r_{ij}$ denotes a semantic relation between $v_i$ and $v_j$ generated from an open-ended vocabulary.

A prompting-based method was used for scene graph generation, bypassing the need for dedicated training. To construct the scene graph, first the graph nodes were instantiated using a list of detected objects from an open-set object detector like Grounding DINO \citep{GDINO} or a VLM. To build the edges, the semantic reasoning capabilities of a VLM are utilized to generate attributes and relationships from an open-vocabulary. Prompting the VLM for all object pairs requires $\mathcal{O}(n^2)$ computational complexity, so instead, following SG-Nav \citep{yin2024sgnav}, the VLM was prompted with the set of objects to jointly generate their attributes and relationships in one shot, reducing the computational cost. The VLM was constrained to strictly output a JSON object, which enables deterministic parsing by the retrieval functions.

\subsection{Reasoning Process}
Once the scene graph $G$ is constructed, the LLM follows an iterative reasoning framework allowing it to utilize the graph effectively for task completion. Due to its single-pass and sequential generation process, chain-of-thought prompting \citep{cot} is heavily dependent on early decisions, allowing initial errors to propagate and cause errors in the final answer \citep{art}. In contrast, the core of SG-CoT is an iterative, multi-turn reasoning process that enables the LLM to actively query the scene graph, ground its thought process, and refine its understanding before committing to an action. We hypothesize that iteratively grounding the LLM's thought process in reality can mitigate accumulation of previous errors and subsequent hallucination. In addition, the LLM is able to pinpoint the root cause of the ambiguity, allowing it to ask a relevant clarification question that will help in disambiguating the situation.

\begin{figure*}[t]
    \centering
    \includegraphics[width=\textwidth]{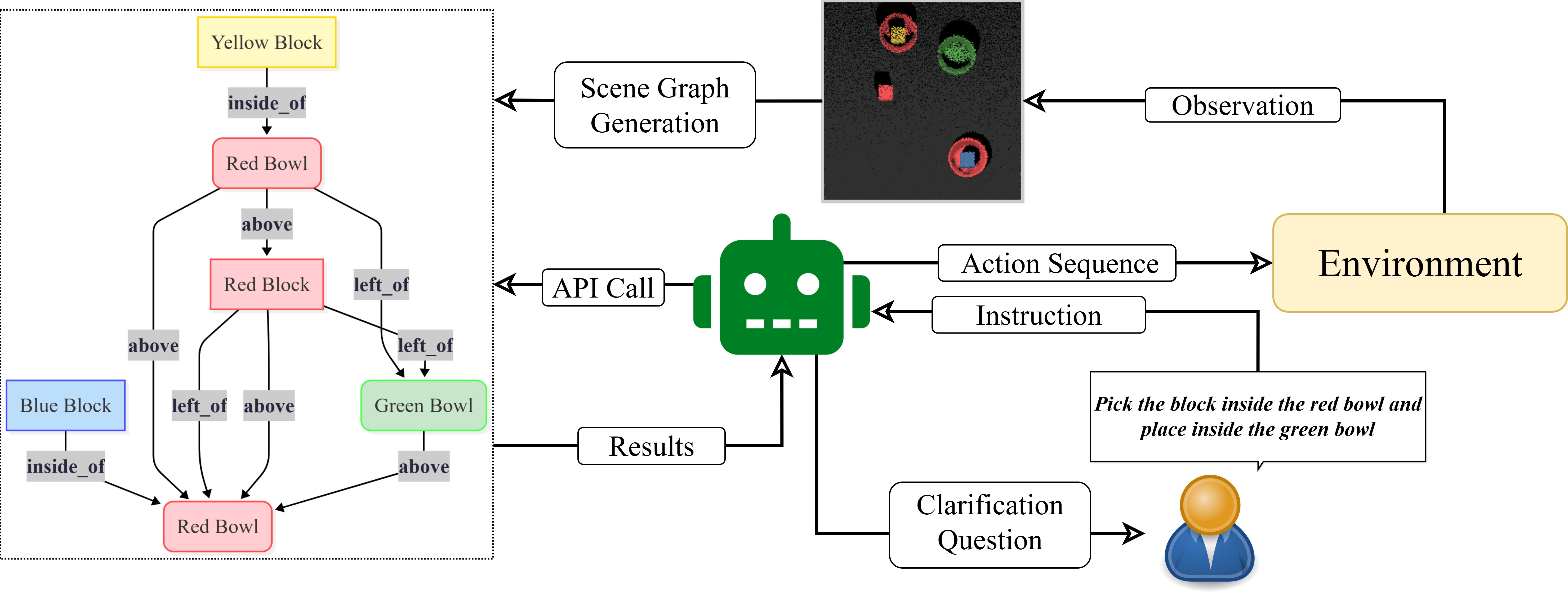} 
    \caption{SG-CoT. The robot receives an instruction from the user and an observation from the environment. The observation is converted to a scene graph representation. During its reasoning process, the robot queries the scene graph via API calls to fetch information about the environment. After multiple rounds of thinking and retrieval, the LLM either performs a set of actions in the environment, or asks a clarification question from the user.}
    \label{fig:sg_full_width}
\end{figure*}

The approach follows a loop of observation, reasoning, function calling and clarification which continues until the LLM has sufficient context to generate a definitive action sequence or a targeted clarification question. This loop ensures that each reasoning step is validated with the scene graph or the user, preventing propagation of hallucinations. The process is initialized by constructing an initial prompt $P_0$, which contains the user instruction $I$, the set of detected objects, and the schema of the scene graph. The schema includes the set of all valid attribute types (e.g., \texttt{color}, \texttt{size}) and edge types (e.g., \texttt{on\_top\_of}, \texttt{inside\_of}), providing the LLM with the necessary vocabulary for querying the graph. A conversation history, $H$, is maintained and is initially set to this prompt, $P_0$.

\begin{algorithm}[t]
\caption{SG-CoT}
\label{alg:sg_cot_reasoning}
\begin{algorithmic}[1]
\State \textbf{Input:} User instruction $I$, Observation $O$, Maximum turns $max\_turns$
\State \textbf{Output:} Action sequence $A$

\Procedure{SG-CoT}{$I, O, max\_turns$}
    \State $X \gets \text{DetectObjects}(O)$
    \State $G \gets \text{SceneGraph}(O,X)$
    \State $P_0 \gets I \oplus X \oplus \text{schema of } G$
    \State $H_0 \gets P_0$
    \State $t \gets 0$
    \State $a_t \gets \varnothing$

    \While{$a_t$ is not an action sequence $A$ \textbf{and} $t < max\_turns$}
        \State $T_t, a_t \gets \text{LLM}(H_t)$
        
        \If{$a_t$ is a function call $f_t$}
            \State $R_t \gets f_t(G)$
            \State $H_{t+1} \gets H_t \oplus T_t \oplus f_t \oplus R_t$
            \State $t \gets t+1$

        \ElsIf{$a_t$ is a clarification question $Q_t$}
            \State $R_t \gets \text{AskUser}(Q_t)$
            \State $H_{t+1} \gets H_t \oplus T_t \oplus Q_t \oplus R_t$
            \State $t \gets t+1$

        \ElsIf{$a_t$ is an action sequence $A$}
            \State \textbf{return} $A$
        \EndIf
    \EndWhile

    \If{$a_t$ is not an action sequence $A$}
        \State $T_t, A \gets \text{LLM}(H_t)$
        \State \textbf{return} $A$
    \EndIf
\EndProcedure
\end{algorithmic}
\end{algorithm}
Following initialization, the process enters an iterative loop where at each turn $t$, the LLM is prompted with the accumulated history $H_t$. It generates a response consisting of two parts. First, a reasoning trace, $T_t$, which details the step-by-step articulation of its reasoning. Secondly, if the LLM concludes that more information is needed it ends its turn by outputting either a clarification question $Q_t$ or a retrieval function $f_t$. if the output is a retrieval function, the system intercepts $f_t$ from the response generated by the LLM. The function queries $G$ and retrieves the relevant nodes or edges. The LLM has access to two different retrieval functions, \texttt{retrieve\_node} and \texttt{retrieve\_edge}. The first function, \texttt{retrieve\_node}, extracts a list of objects from the graph based on some condition. It can be either a direct object \texttt{name} or by a specific attribute key-value pair. For example, a query like \texttt{retrieve\_node(attr\_key='color', attr\_val= 'red')} would return the names of all nodes in the scene that have the color red.

The second function, \texttt{retrieve\_edge}, can be used to query the spatial and semantic relationships between nodes. It accepts optional arguments for the \texttt{source} node, the \texttt{target} node, and the \texttt{relation} type. These parameters can be used individually or in combination to filter the scene's edge list. For example, \texttt{retrieve\_edge(source='blue bowl', relation='above')} would return a list of strings detailing all the objects the `blue bowl' is `above', such as ``the blue bowl is above the red block".

After the function call is made, it returns the results $R_t$. The conversation history is updated according to the rule $H_{t+1} = H_t \oplus (T_t \oplus f_t \oplus R_t)$, where $\oplus$ denotes concatenation. The LLM is then prompted with the augmented history, $H_{t+1}$ through which it can review its previous thought process with the retrieval results and make inferences about the environment.

If at any stage the LLM concludes that the scene graph cannot provide sufficient information for it to execute a definitive action sequence, it outputs a clarification question $Q_t$. this question is presented to the user who responds with its response $R_t$. the question and the response are updated to the conversation history using the same rule as the retrieval functions.

The loop of reasoning continues until the termination condition is met. This occurs when the LLM's thought process in $T_t$ concludes that it has sufficient information to take a definitive action. Alternatively, the process terminates if the model reaches a predefined limit of \texttt{max\_turns} rounds of thought, after which it is prompted to conclude its reasoning. At this point, instead of calling another retrieval function, it generates a final action sequence, $A = (a_1, a_2, \dots, a_k)$.

\section{Experimentation}\label{sec4}
Two simulated tabletop experiments were conducted to evaluate the effectiveness of SG-CoT in both single-agent and partially observable multi-agent environments. For both experiments, the method was compared against three existing baselines: an uncertainty-estimation based method, CLARA  \citep{CLARA}; a feedback mechanism based approach, Inner Monologue  \citep{huang2022inner}; and a robust prompting technique, ProgPrompt  \citep{ProgPrompt}, extended to be able to ask for clarification as an action. The experiments were conducted using \texttt{gemini-2.5-flash}  \citep{gemini} and \texttt{Qwen3-VL-2B-Instruct}  \citep{qwen} as the VLM. Objects were also detected using the same model. Through the experiments, we aim to answer the following research questions: (1) How well does SG-CoT perform under various kinds of ambiguities? (2) Given an ambiguous environment, can it ask a relevant clarification question? (3) Can different robots communicate with each other for disambiguation in partially observed environments using SG-CoT?

\subsection{Evaluation Metrics}
Each baseline was evaluated over the following two metrics.
\begin{itemize}
    \item \textbf{Ambiguity Detection Accuracy (ADA):} This metric defines how often can the model correctly identify an  ambiguous situation. For unambiguous situations \textit{ADA} is 1 only if the correct action sequence is executed and 0 otherwise. For ambiguous, \textit{ADA} is 1 when the robot asks for clarification during its reasoning, and 0 otherwise.
    
    \item \textbf{Success Rate (SR):} This metric determines if the task was finally completed. \textit{SR} is 1 when the final environment state matches the user request, and 0 otherwise. Furthermore, \textit{SR} inherently evaluates the correctness of the clarification question, since an inaccurate question will fail to resolve the ambiguity, ultimately resulting in an \textit{SR} of 0.
\end{itemize}

\subsection{Single-Agent}
The first experiment is conducted over a single-agent environment, evaluating the method over the first three ambiguity types, multiplicity, absence and underspecification. In this section, the experimental setup is described first, followed by the results.

\begin{table}[t]
\centering
\caption{Instructions for single-agent experiment}
\label{t1}
\renewcommand{\arraystretch}{1.5}
\begin{tabularx}{\linewidth}{>{\hsize=.6\hsize}X >{\hsize=1.4\hsize}X}
\hline
\textbf{Task Types} & \textbf{Instruction Template} \\
\hline

\textbf{Basic}  & Pick up [TARGET PICK] and place on [TARGET PLACE] \\

\textbf{Spatial}  & Pick up the object to the [DIRECTION] of the [REFERENCE] and place on [TARGET PLACE] \\

\textbf{Long Horizon} & Place all the blocks over the bowls that match their color \\

\textbf{Underspecified} & Place [VAGUE] over the [TARGET PLACE] \\

\hline
\end{tabularx}
\end{table}

\begin{table*}[b]
\centering
\normalsize
\caption{Performance assessment: ADA and SR across ambiguous scenarios in single-agent environments. Values are reported as ADA/SR.}
\label{tab:t2}
\begin{tabular}{lcccc}
\hline
\textbf{Method} &
\textbf{Multiplicity} &
\textbf{Absence} &
\textbf{Underspecified} &
\textbf{Overall} \\
\hline

\multicolumn{5}{c}{\texttt{\textbf{Qwen3-VL-2B}}} \\
\hline
CLARA\citep{CLARA} &
0.38/0.35 &
0.35/0.35 &
0.63/0.52 &
0.44/0.40 \\

ProgPrompt\citep{ProgPrompt} &
0.25/0.23 &
0.33/0.28 &
0.58/0.47 &
0.37/0.32 \\

InnerMono\citep{huang2022inner} &
0.65/\textbf{0.53} &
0.67/0.52 &
0.75/0.62 &
0.65/0.55 \\

SG-CoT &
\textbf{0.71}/0.52 &
\textbf{0.90}/\textbf{0.80} &
\textbf{0.77}/\textbf{0.67} &
\textbf{0.72}/\textbf{0.66} \\
\hline

SG-CoT (0 iter) &
0.34/0.32 &
0.32/0.26 &
0.44/0.28 &
0.38/0.28 \\

SG-CoT (1 iter) &
0.45/0.35 &
0.60/0.42 &
0.69/0.62 &
0.54/0.46 \\

SG-CoT (w/o E) &
0.52/0.37 &
0.69/0.41 &
0.75/0.52 &
0.55/0.43 \\
\hline

\multicolumn{5}{c}{\textbf{\texttt{gemini-2.5-flash}}} \\
\hline
CLARA\citep{CLARA} &
0.48/0.47 &
0.37/0.33 &
0.70/0.60 &
0.52/0.49 \\

ProgPrompt\citep{ProgPrompt} &
0.26/0.24 &
0.48/0.31 &
0.79/0.77 &
0.55/0.45 \\

InnerMono\citep{huang2022inner} &
0.70/\textbf{0.62} &
0.81/0.48 &
0.95/0.93 &
0.76/0.68 \\

SG-CoT &
\textbf{0.77}/0.51 &
\textbf{0.96}/\textbf{0.86} &
\textbf{0.99}/\textbf{0.98} &
\textbf{0.80}/\textbf{0.78} \\
\hline

SG-CoT (0 iter) &
0.32/0.28 &
0.44/0.36 &
0.68/0.52 &
0.48/0.39 \\

SG-CoT (1 iter) &
0.52/0.40 &
0.71/0.45 &
0.82/0.75 &
0.63/0.53 \\

SG-CoT (w/o E) &
0.59/0.32 &
0.84/0.49 &
0.87/0.54 &
0.63/0.44 \\
\hline
\end{tabular}
\end{table*}

\textit{Experimental Setup:} The experimental setup is based on simulated tabletop environments adapted from SayCan  \citep{saycan2022arxiv}, a PyBullet-based environment with a UR5e and 2-finger gripper. The baselines are evaluated over three different tasks defined in the environment, with their corresponding instruction templates shown in Table~\ref{t1}. For the basic task configuration, the pick, place, and distractor objects were randomly sampled from a predefined object list and place them within the environment. For the long-horizon setup, a subset of colors was randomly selected from the available color set and instantiate blocks and bowls corresponding to the sampled colors. For the spatial configuration, first a reference object is randomly selected and placed it at the center of the workspace. A spatial relation is then chosen by sampling a direction (\texttt{left}, \texttt{right}, \texttt{above}, or \texttt{below}) and positioning the target object accordingly relative to the reference. Finally, distractor objects are placed around the reference object, excluding the direction chosen for the target placement.

To incorporate the ambiguity types described earlier, the environment and instructions were modified as follows. Underspecification was introduced by replacing the target object in the instruction with a term drawn from a curated dictionary of vague words. For absence, the target object was removed, and for multiplicity the target object was duplicated. In total, 400 trials were conducted with 100 randomized configurations per ambiguity category, and 100 clear scenarios. In each trial, the LLM received the current observation and the task instruction. The action space consisted of two high-level actions: \texttt{pick\_and\_place} and \texttt{ask}, which together define a planning interface over the environment state. The action \texttt{pick\_and\_place} expects two arguments, the object to be picked and the object where it should be placed, and is executed on the precondition that no object is currently held, resulting in the object being placed inside or on the specified receptacle. The action \texttt{ask} expects a question string and an ambiguity tag defining the type of ambiguity, and is invoked when the observation does not allow a unique grounding of the instruction. If the tag predicted by the LLM matches the ground truth ambiguity type, the LLM is provided with the missing target. However, if the predicted tag does not match the ground truth, the trial is immediately terminated, resulting in an \textit{SR} of 0 for the trial, as the LLM was not able to identify the source of ambiguity. The goal condition for each trial is satisfied when the intended target object is successfully placed in the specified container or spatial location.

\textit{Results:} As shown in Table~\ref{tab:t2}, instruction-level ambiguity was the most manageable for all previous approaches with an average \textit{ADA} of 77\% compared to 56.6\% for ambiguities at the environment-level as they are not explicitly visible from the instruction alone. SG-CoT maintained high performance across both settings, achieving \textit{ADA} of 99.0\% and 77\% under underspecified instructions and 86.5\% and 80.5\% for environment-level ambiguities with \texttt{gemini-2.5-flash} and \texttt{Qwen3-VL-2B-Instruct} respectively. The notable increase in SR by 10\% with \texttt{gemini-2.5-flash} and 11\% with \texttt{Qwen3-VL-2B-Instruct} obtained by SG-CoT over the previous best Inner Monologue, supports the hypothesis that selectively retrieving only the environment elements relevant to task completion enables the model to formulate accurate clarification questions. There is a performance drop in multiplicity scenarios as compared to the other cases, as it still allows the LLM to take an action that matches the user request, unlike absence where picking up the target block would be inherently invalid. For example, if two valid red blocks are present, the model may arbitrarily pick one to complete the user instruction, ignoring the ambiguity. In contrast, if the scene graph does not return a block, the ambiguity is relatively explicit and easier for the model to detect. The evaluation metrics were consistently higher for \texttt{gemini-2.5-flash} compared to \texttt{Qwen3-VL-2B-Instruct}, suggesting that performance scales positively with model capacity. Overall, SG-CoT achieves consistently superior performance across both evaluation metrics and LLMs compared to prior approaches.

\textit{Ablation:} An ablation study was conducted to analyze the contribution of individual components in the framework. In the first ablation, instead of iteratively retrieving relevant parts of the scene graph, the model is provided with the entire scene graph directly, based on which it must predict and action sequence. This results in an overall drop of 0.34/0.38 (ADA/SR) for \texttt{Qwen3-VL-2B-Instruct}, and 0.32/0.39 for \texttt{gemini-2.5-flash}, demonstrating the role of iterative reasoning. Restricting the model to a single retrieval step before removing the scene graph access further leads to a lower decrease of 0.18/0.20 for \texttt{Qwen3-VL-2B-Instruct}, and 0.17/0.25 for \texttt{gemini-2.5-flash}. For the final ablation, the scene graph is built without the edges, resulting in reductions of 0.17/0.23 for \texttt{Qwen3-VL-2B-Instruct}, and 0.17/0.34 for \texttt{gemini-2.5-flash}, showcasing the role of scene graph accuracy.

\begin{table}[t]
\centering
\renewcommand{\arraystretch}{1.15}
\caption{Computational cost analysis across methods. We report average LLM calls per episode and average input/output tokens per call.}
\label{tab:cost}
\begin{tabular}{lccc}
\hline
Method & Calls/Ep & Avg In/Call & Avg Out/Call\\
\hline
CLARA      & 10.36 & 354.96  & 21.47 \\
ProgPrompt & 1.00  & 410.81  & 448.36 \\
InnerMono  & 2.24  & 128.42  & 35.79 \\
SG-CoT     & 4.24  & 1141.72 & 191.28 \\
\hline
\end{tabular}
\end{table}

\textit{Computational Cost and Scalability:} As shown in Table~\ref{tab:cost}, due to the history concatenation and the additional scene graph generation step, SG-CoT does require higher token usage and LLM calls. The latency also scales approximately linearly with the total number of tokens processed and the number of calls. However, increasing the total number of objects in the environment does not proportionally increase the number of reasoning iterations, since scene graph generation is performed in a single pass and subsequent retrieval depth depends upon the size of the user request, as the model only retrieves parts of the graph relevant to the user request, supporting scalability with increase in environment size.

\subsection{Partial Observability}
This section entails the setup and the performance of the proposed method over the final ambiguity type caused by partial observability in multi-agent environments, with a focus on inter-agent communication for resolving observation-level ambiguity.

\begin{table*}[b]
\centering
\caption{Performance assessment: SR in partially observed environments. Results are reported using Qwen3-VL-2B and gemini-2.5-flash.}
\label{tab:multi}
\begin{tabular}{lcc}
\hline
\textbf{Method} & \texttt{\textbf{Qwen3-VL-2B}} & \texttt{\textbf{gemini-2.5-flash}} \\
\hline
CLARA      & 0.39 & 0.45 \\
ProgPrompt & 0.44 & 0.59 \\
InnerMono  & 0.45 & 0.48 \\
SG-CoT     & \textbf{0.59} & \textbf{0.75} \\
\hline
\end{tabular}
\end{table*}

\textit{Experimental Setup:} To evaluate the performance of SG-CoT in handling observation-level ambiguity within multi-agent systems, the same environment was utilized as before but the tasks were inspired from the LEMMA  \citep{Lemma} benchmark. The environment consists of two cooperative robotic agents, each operating under partial observability with access to a distinct individual workspace as well as a shared workspace. Each robot can perceive only the objects within its own and the shared workspace, thereby requiring communication with the other agent to obtain missing observations needed for task completion. As illustrated in Fig.~\ref{fig:amb}, the work space was divided into three regions, red being the workspace for first robot, blue for the second robot, and the green area serves as the shared workspace. The input of each robot was occluded to introduce the partial observability.

Ambiguity via partial observability is injected into these tasks by transporting the target block to be picked up out of the working space of the robot. In this case, the robot must communicate with the other robot to complete the task. To achieve this, the action space was expanded by introducing a new clarification action, \texttt{ask\_robot}. This action enables an agent to query the other robot regarding its local observations. The input argument to the action is a string that is passed to the other robot. If the target objects required for task completion are not present in the robot's view the robot must ask for observations of the other robot and use the shared workspace for transfer of objects. If the target place is missing form the robots view, the robot must place the block in the shared space to allow the other robot to to allow the other robot to pick it up. Since there is only one ambiguity type, only \textit{SR} was calculated for this experiment, as correct detection of ambiguity in this case always led to the model completing its task. In total 200 trials were conducted each with random configurations.

\textit{Results:} As shown by Table.~\ref{tab:multi}, SG-CoT achieves \textit{SR} of 75\% with \texttt{gemini-2.5-flash} and 59\% with \texttt{Qwen3-VL-2B-Instruct} over observation-level ambiguity, corresponding to improvements of 16\% and 15\% respectively over the next best baseline, ProgPrompt, showing its effectiveness in communication-driven reasoning under partial observability. Since prior approaches depend on simpler environmental representations, they can hallucinate details about the environment and resort to executing an action instead of asking a question. SG-CoT forces an LLM to query parts of the scene graph for details of the environment, preventing it from making up new details and rely solely on the returned results for its reasoning.

\section{Conclusion}\label{sec5}
In this paper, Scene Graph-Chain-of-Thought (SG-CoT) was proposed, a framework that allows large language model (LLM) based robotic planners to iteratively query and reason over a scene graph representation of the environment in order to make decisions. This framework allows the LLM to be resistant to ambiguity by grounding its reasoning in reality, shown by a 4\% enhancement in overall success rate in single-agent environments and 15\% in multi-agent settings. The method also allows the LLM to pinpoint the reason of ambiguity and ask a question required for disambiguation, leading to a minimum 10\% improvement in success rate. However, there are some limitations to SG-CoT. The current evaluation is limited to simulation-based experiments, and no real-world experiments were performed where perception noise and execution
uncertainty are inherently present. In addition, the method relies on a vision–language model (VLM) to infer spatial and semantic relationships in a single step, which can hallucinate with real-world objects and introduce incorrect edges into the scene graph, potentially hampering the reasoning process. Finally, the token usage and inference latency increase with the number of ambiguities present in the scene, which may limit scalability for more complex instructions . In the future, the framework could be extended to embodied AI tasks such as vision–language navigation in real-world indoor environments.

\begin{appendices}

% ==========================================
% SECTION A: SYSTEM PROMPTS
% ==========================================
\section{Prompts}\label{sec:system_prompts}

This section details the exact prompts used for both scene graph generation and step-by-step reasoning.

\subsection{Scene Graph Generation Prompt}\label{subsec:prompt_sg}

\begin{tcolorbox}[
enhanced,
breakable,
colback=black!5,
colframe=black!70,
title=Scene Graph Generation Prompt,
coltitle=white,
colbacktitle=black!80,
fonttitle=\bfseries,
fontupper=\ttfamily\normalsize,
boxrule=0.8pt,
arc=2mm,
left=2mm,
right=2mm,
top=1mm,
bottom=1mm
]
You are a scene graph generator. Your task is to generate a scene graph representation of a given tabletop scene, identifying relationships between all the objects presented to you (if any).

The graph must have the following structure:

Nodes: The nodes must be represented by objects. You will already be provided with the object list.

Edges: These should represent the relationships between each pair of objects. These relationships can include spatial directions (left\_of, above, below, on\_top\_of, etc.) or semantic relations (joined\_with, inside\_of, etc.) to properly model how each object is related to the others.

Attributes: Each node must have an attribute of the object like color, type etc.

Input: You will be given the image of a tabletop scene as input, as well as the object list over the table to label the nodes.

Output: Must strictly be a JSON object with the following schema:

\{"nodes": [\{"id": $i$, "attributes": \{a\_1, a\_2, \dots, a\_k\}\}], "edges": [\{"source": $i$, "target": $j$, "relation": $r$\}]\}
\end{tcolorbox}

\subsection{Reasoning Prompt}
\label{subsec:prompt_reasoning}

\begin{tcolorbox}[
enhanced,
breakable,
colback=black!5,
colframe=black!70,
title=Reasoning Prompt,
coltitle=white,
colbacktitle=black!80,
fonttitle=\bfseries,
fontupper=\ttfamily\normalsize,
boxrule=0.8pt,
arc=2mm,
left=2mm,
right=2mm,
top=1mm,
bottom=1mm
]
You are a Reasoning agent for a robot. Your will be given an instruction and a scene graph of the environment based on which you have to choose an action sequence or ask a relevant question.

You do not have access to the entire graph, rather you will have to use retrieval functions to get access to relevant areas of the graph to make a decision.

Your decisions must strictly be grounded in the attributes of the nodes and the results of the retrieval functions.

If you decide to call a retrieval function, ask a clarification question, or indicate that an object is absent, you must end your output there. The function will then be called and you will be presented the output in a separate call.

You have access to two retrieval functions:

1. retrieve\_edge(scene, source="node of the source of the edge", target="node of the destination of the edge", relation="type of edge")

2. retrieve\_node(scene, name="object name", attr\_key="attribute name", attr\_val="value")

"scene" is the only necessary argument; all others can be used as filters (if required) to access different parts of the graph.

You also have access to two question functions to be asked to the user in case an ambiguity arises:

3. ask\_multiplicity(question="question with multiple options")

4. ask\_absence(question="question indicating object not found")

When you need information, request it with the following format:

THOUGHT: the thinking based on the conversation history, result of the retrieval, or the instruction.

REQUEST: retrieve\_node()

REQUEST: retrieve\_edge()

REQUEST: ask\_multiplicity("question string")

REQUEST: ask\_absence("question string")

FINAL\_DECISIONS: [
\{"action": "pick\_and\_place",
"pick\_object": "[object\_name]",
"place\_object": "[object\_name]"\}
]

If clarification from the user is required, use ask\_multiplicity() and end your output there.

If a required object is not found, use ask\_absence() and end your output there.

Always return FINAL\_DECISIONS as a JSON list of action objects. Even if there is only one action, it must be a single-element list.

Do not include any extra commentary after the FINAL\_DECISIONS block.
\end{tcolorbox}
% ==========================================
% SECTION B: REASONING EXAMPLES
% ==========================================
\section{LLM Reasoning Examples}
\label{sec:reasoning_examples}

The following examples demonstrate the step-by-step reasoning process of the LLM for all cases of ambiguities.

\subsection{Example 1: Multiplicity Ambiguity}
\label{subsec:mul_ambiguity}

\begin{tcolorbox}[
enhanced,
breakable,
colback=lightgray,
colframe=black,
boxrule=0.4pt,
arc=2pt,
left=4pt,
right=4pt,
top=4pt,
bottom=4pt,
listing only,
before upper=\raggedright
]
{\color{user} User: Pick the block inside the red bowl and place it inside the green bowl.}

{\color{llmthink} LLM: The user wants me to place the block inside the red bowl into the green bowl. First, I will identify all red bowls in the scene.}

{\color{llmresp} LLM: retrieve\_node(type=`bowl', color=`red')}

{\color{sgcolor} SG: [`red bowl 1', `red bowl 2']}

{\color{llmthink} LLM: There are two red bowls. I should next determine which objects are inside each of them so I can identify the relevant block.}

{\color{llmresp} LLM: retrieve\_edge(type=`inside\_of', target=`red bowl 1')}

{\color{llmresp} LLM: retrieve\_edge(type=`inside\_of', target=`red bowl 2')}

{\color{sgcolor} SG: Yellow block is inside\_of the red bowl 1, Blue block is inside\_of the red bowl 2}

{\color{llmthink} LLM: There are two red bowls, each with a different block, making the user’s command ambiguous. I should ask for clarification.}

{\color{llmresp} LLM: ask\_multiplicity("Did you mean the yellow block or the blue block?")}

{\color{user} User: The blue block.}

{\color{llmthink} LLM: The user clarified that they mean the blue block. I now need to identify the green bowl that will serve as the destination.}

{\color{llmresp} LLM: retrieve\_node(type=`bowl', color=`green')}

{\color{sgcolor} SG: [`green bowl 1']}

{\color{llmthink} LLM: There is exactly one green bowl in the scene, so the destination is unambiguous. I can now execute the requested action.}

{\color{llmresp} LLM: pick\_and\_place(object=`blue block', destination=`green bowl 1')}
\end{tcolorbox}

\subsection{Example 2: Absence Ambiguity}
\label{subsec:abs_ambiguity}

\begin{tcolorbox}[
enhanced,
breakable,
colback=lightgray,
colframe=black,
boxrule=0.4pt,
arc=2pt,
left=4pt,
right=4pt,
top=4pt,
bottom=4pt,
listing only,
before upper=\raggedright
]
{\color{user} User: Pick up the blue block and place it inside the purple bowl.}

{\color{llmthink} LLM: The task specifies the blue block and a destination purple bowl. I should first determine whether the destination exists in the scene.}

{\color{llmresp} LLM: retrieve\_node(type=`bowl', color=`purple')}

{\color{sgcolor} SG: []}

{\color{llmthink} LLM: No purple bowl is present in the scene. The requested destination is absent, so I cannot complete the task as specified. I should ask the user for an alternative destination.}

{\color{llmresp} LLM: ask\_absence("I cannot find a purple bowl in the scene. Where would you like me to place the blue block instead?")}

{\color{user} User: Place it inside the green bowl.}

{\color{llmthink} LLM: The user has provided an alternative destination. I should verify that the green bowl exists.}

{\color{llmresp} LLM: retrieve\_node(type=`bowl', color=`green')}

{\color{sgcolor} SG: [`green bowl 1']}

{\color{llmthink} LLM: The green bowl exists and is uniquely identified. I can now execute the task.}

{\color{llmresp} LLM: pick\_and\_place(object=`blue block', destination=`green bowl 1')}
\end{tcolorbox}

\subsection{Example 3: Underspecified Ambiguity}
\label{subsec:ex_underspecified}

\begin{tcolorbox}[
enhanced,
breakable,
colback=lightgray,
colframe=black,
boxrule=0.4pt,
arc=2pt,
left=4pt,
right=4pt,
top=4pt,
bottom=4pt,
listing only,
before upper=\raggedright
]
{\color{user} User: Can you move the red block away?}

{\color{llmthink} LLM: The instruction specifies the object (red block) but does not specify the target location. I need clarification about where the block should be placed.}

{\color{llmresp} LLM: ask\_underspecified(``Where exactly would you like me to place the red block?")}

{\color{user} User: Place it inside the green bowl.}

{\color{llmthink} LLM: The user has specified the destination as the green bowl. I should verify that there is a unique green bowl in the scene before executing the action.}

{\color{llmresp} LLM: retrieve\_node(type=`bowl', color=`green')}

{\color{sgcolor} SG: [`green bowl 1']}

{\color{llmthink} LLM: There is exactly one green bowl in the scene, so the destination is unambiguous. I can now execute the requested action.}

{\color{llmresp} LLM: pick\_and\_place(object=`red block', destination=`green bowl 1')}
\end{tcolorbox}

\subsection{Example 4: Multi-Agent Collaboration}
\label{subsec:ex_collaboration}

\begin{tcolorbox}[
enhanced,
breakable,
colback=lightgray,
colframe=black,
boxrule=0.4pt,
arc=2pt,
left=4pt,
right=4pt,
top=4pt,
bottom=4pt,
listing only,
before upper=\raggedright
]
{\color{user} User: Move the blue block into the yellow bowl.}

{\color{llmthink} Robot 1: The task requires moving the blue block into the yellow bowl. I will first check whether the blue block is visible in my workspace or the shared workspace.}

{\color{llmresp} Robot 1: retrieve\_node(type=`block', color=`blue')}

{\color{sgcolor} SG: []}

{\color{llmthink} Robot 1: The blue block is not visible in my observation. Since the environment is partially observable, the block may be located in Robot 2's workspace. I should query Robot 2.}

{\color{llmresp} Robot 1: ask\_robot(``Can you see a blue block in your workspace?")}

{\color{llmthink} Robot 2: Robot 1 cannot observe the blue block. I should check whether it is present in my workspace and, if found, transfer it to the shared workspace.}

{\color{llmresp} Robot 2: retrieve\_node(type=`block', color=`blue')}

{\color{sgcolor} SG: [`blue block 1']}

{\color{llmthink} Robot 2: I can see the blue block. I should transfer the block through the shared workspace.}

{\color{llmresp} Robot 2: pick\_and\_place(object=`blue block 1', destination=`shared\_space')}

{\color{llmthink} Robot 1: The blue block has been transferred to the shared workspace. I now need to determine whether the target yellow bowl is visible.}

{\color{llmresp} Robot 1: retrieve\_node(type=`bowl', color=`yellow')}

{\color{sgcolor} SG: [`yellow bowl 1']}

{\color{llmthink} Robot 1: The yellow bowl is visible and uniquely identified. I can complete the task.}

{\color{llmresp} Robot 1: pick\_and\_place(object=`blue block 1', destination=`yellow bowl 1')}
\end{tcolorbox}

\end{appendices}

\bibliography{sn-bibliography}% common bib file

@inproceedings{knowno,
    title={Robots That Ask For Help: Uncertainty Alignment for Large Language Model Planners},
    author={Ren, Allen Z. and Dixit, Anushri and Bodrova, Alexandra and Singh, Sumeet and Tu, Stephen and Brown, Noah and Xu, Peng and Takayama, Leila and Xia, Fei and Varley, Jake and Xu, Zhenjia and Sadigh, Dorsa and Zeng, Andy and Majumdar, Anirudha},
    booktitle={Proceedings of the Conference on Robot Learning (CoRL)},
    year={2023}
}

@article{CLARA,
          journal={IEEE Robotics and Automation Letters}, 
          title={CLARA: Classifying and Disambiguating User Commands for Reliable Interactive Robotic Agents}, 
          author={Jeongeun Park and Seungwon Lim and Joonhyung Lee and Sangbeom Park and Minsuk Chang and Youngjae Yu and Sungjoon Choi},
          year={2024},
          volume={9},
          number={2},
          pages={1059-1066},
          doi={10.1109/LRA.2023.3338514}}

@inproceedings{chen2024mapgpt,
  title={MapGPT: Map-Guided Prompting with Adaptive Path Planning for Vision-and-Language Navigation},
  author={Chen, Jiaqi and Lin, Bingqian and Xu, Ran and Chai, Zhenhua and Liang, Xiaodan and Wong, Kwan-Yee~K.},
  booktitle = "Proceedings of the 62nd Annual Meeting of the Association for Computational Linguistics",
  year={2024}
}

@article{zhou2023navgpt,
  title={NavGPT: Explicit Reasoning in Vision-and-Language Navigation with Large Language Models},
  author={Zhou, Gengze and Hong, Yicong and Wu, Qi},
  journal={arXiv preprint arXiv:2305.16986},
  year={2023}
}

@inproceedings{huang2022inner,
    title={Inner Monologue: Embodied Reasoning through Planning with Language Models},
    author={Wenlong Huang and Fei Xia and Ted Xiao and Harris Chan and Jacky Liang and Pete Florence and Andy Zeng and Jonathan Tompson and Igor Mordatch and Yevgen Chebotar and Pierre Sermanet and Noah Brown and Tomas Jackson and Linda Luu and Sergey Levine and Karol Hausman and Brian Ichter},
    booktitle={arXiv preprint arXiv:2207.05608},
    year={2022}
}

@inproceedings{ivanova-etal-2025-ambik,
    title = "{A}mbi{K}: Dataset of Ambiguous Tasks in Kitchen Environment",
    author = "Ivanova, Anastasia  and
      Eva, Bakaeva  and
      Volovikova, Zoya  and
      Kovalev, Alexey  and
      Panov, Aleksandr",
    editor = "Che, Wanxiang  and
      Nabende, Joyce  and
      Shutova, Ekaterina  and
      Pilehvar, Mohammad Taher",
    booktitle = "Proceedings of the 63rd Annual Meeting of the Association for Computational Linguistics (Volume 1: Long Papers)",
    month = jul,
    year = "2025",
    address = "Vienna, Austria",
    publisher = "Association for Computational Linguistics",
    url = "https://aclanthology.org/2025.acl-long.1593/",
    doi = "10.18653/v1/2025.acl-long.1593",
    pages = "33216--33241",
    ISBN = "979-8-89176-251-0"
}

@inproceedings{saycan2022arxiv,
    title={Do As I Can and Not As I Say: Grounding Language in Robotic Affordances},
    author={Michael Ahn and Anthony Brohan and Noah Brown and Yevgen Chebotar and Omar Cortes and Byron David and Chelsea Finn and Chuyuan Fu and Keerthana Gopalakrishnan and Karol Hausman and Alex Herzog and Daniel Ho and Jasmine Hsu and Julian Ibarz and Brian Ichter and Alex Irpan and Eric Jang and Rosario Jauregui Ruano and Kyle Jeffrey and Sally Jesmonth and Nikhil Joshi and Ryan Julian and Dmitry Kalashnikov and Yuheng Kuang and Kuang-Huei Lee and Sergey Levine and Yao Lu and Linda Luu and Carolina Parada and Peter Pastor and Jornell Quiambao and Kanishka Rao and Jarek Rettinghouse and Diego Reyes and Pierre Sermanet and Nicolas Sievers and Clayton Tan and Alexander Toshev and Vincent Vanhoucke and Fei Xia and Ted Xiao and Peng Xu and Sichun Xu and Mengyuan Yan and Andy Zeng},
    booktitle={arXiv preprint arXiv:2204.01691},
    year={2022}
}

@inproceedings{rt22023arxiv,
    title={RT-2: Vision-Language-Action Models Transfer Web Knowledge to Robotic Control},
    author={Anthony Brohan and Noah Brown and Justice Carbajal and Yevgen Chebotar and Xi Chen and Krzysztof Choromanski and Tianli Ding and Danny Driess and Avinava Dubey and Chelsea Finn and Pete Florence and Chuyuan Fu and Montse Gonzalez Arenas and Keerthana Gopalakrishnan and Kehang Han and Karol Hausman and Alex Herzog and Jasmine Hsu and Brian Ichter and Alex Irpan and Nikhil Joshi and Ryan Julian and Dmitry Kalashnikov and Yuheng Kuang and Isabel Leal  and Lisa Lee and Tsang-Wei Edward Lee and Sergey Levine and Yao Lu and Henryk Michalewski and Igor Mordatch and Karl Pertsch and Kanishka Rao and Krista Reymann and Michael Ryoo and Grecia Salazar and Pannag Sanketi and Pierre Sermanet and Jaspiar Singh and Anikait Singh and Radu Soricut and Huong Tran and Vincent Vanhoucke and Quan Vuong and Ayzaan Wahid and Stefan Welker and Paul Wohlhart and  Jialin Wu and Fei Xia and Ted Xiao and Peng Xu and Sichun Xu and Tianhe Yu and Brianna Zitkovich},
    booktitle={arXiv preprint arXiv:2307.15818},
    year={2023}
}

@inproceedings{kim-etal-2024-aligning,
    title = "Aligning Language Models to Explicitly Handle Ambiguity",
    author = "Kim, Hyuhng Joon  and
      Kim, Youna  and
      Park, Cheonbok  and
      Kim, Junyeob  and
      Park, Choonghyun  and
      Yoo, Kang Min  and
      Lee, Sang-goo  and
      Kim, Taeuk",
    editor = "Al-Onaizan, Yaser  and
      Bansal, Mohit  and
      Chen, Yun-Nung",
    booktitle = "Proceedings of the 2024 Conference on Empirical Methods in Natural Language Processing",
    month = nov,
    year = "2024",
    address = "Miami, Florida, USA",
    publisher = "Association for Computational Linguistics",
    url = "https://aclanthology.org/2024.emnlp-main.119/",
    doi = "10.18653/v1/2024.emnlp-main.119",
    pages = "1989--2007"
}

@article{gu2021open,
  title={Open-Vocabulary Detection via Vision and Language Knowledge Distillation},
  author={Gu, Xiuye and Lin, Tsung-Yi and Kuo, Weicheng and Cui, Yin},
  journal={arXiv preprint arXiv:2104.13921},
  year={2021}
}

@inproceedings{li2022blip,
      title={BLIP: Bootstrapping Language-Image Pre-training for Unified Vision-Language Understanding and Generation}, 
      author={Junnan Li and Dongxu Li and Caiming Xiong and Steven Hoi},
      year={2022},
      booktitle={ICML},
}

@INPROCEEDINGS{cap,
  author={Liang, Jacky and Huang, Wenlong and Xia, Fei and Xu, Peng and Hausman, Karol and Ichter, Brian and Florence, Pete and Zeng, Andy},
  booktitle={2023 IEEE International Conference on Robotics and Automation (ICRA)}, 
  title={Code as Policies: Language Model Programs for Embodied Control}, 
  year={2023},
  volume={},
  number={},
  pages={9493-9500},
  doi={10.1109/ICRA48891.2023.10160591}}

@INPROCEEDINGS{ProgPrompt,
  author={Singh, Ishika and Blukis, Valts and Mousavian, Arsalan and Goyal, Ankit and Xu, Danfei and Tremblay, Jonathan and Fox, Dieter and Thomason, Jesse and Garg, Animesh},
  booktitle={2023 IEEE International Conference on Robotics and Automation (ICRA)}, 
  title={ProgPrompt: Generating Situated Robot Task Plans using Large Language Models}, 
  year={2023},
  volume={},
  number={},
  pages={11523-11530},
  doi={10.1109/ICRA48891.2023.10161317}}

@InProceedings{song2023llmplanner,
  author    = {Song, Chan Hee and Wu, Jiaman and Washington, Clayton and Sadler, Brian M. and Chao, Wei-Lun and Su, Yu},
  title     = {LLM-Planner: Few-Shot Grounded Planning for Embodied Agents with Large Language Models},
  booktitle = {Proceedings of the IEEE/CVF International Conference on Computer Vision (ICCV)},
  month     = {October},
  year      = {2023},
}

@misc{CLAM,
      title={CLAM: Selective Clarification for Ambiguous Questions with Generative Language Models}, 
      author={Lorenz Kuhn and Yarin Gal and Sebastian Farquhar},
      year={2023},
      eprint={2212.07769},
      archivePrefix={arXiv},
      primaryClass={cs.CL},
      url={https://arxiv.org/abs/2212.07769}, 
}

@misc{intent-sim,
      title={Clarify When Necessary: Resolving Ambiguity Through Interaction with LMs}, 
      author={Michael J. Q. Zhang and Eunsol Choi},
      year={2023},
      eprint={2311.09469},
      archivePrefix={arXiv},
      primaryClass={cs.CL},
      url={https://arxiv.org/abs/2311.09469}, 
}

@inproceedings{ECLAIR,
author = {Murzaku, John and Liu, Zifan and Muppala, Vaishnavi and Tanjim, Md Mehrab and Chen, Xiang and Li, Yunyao},
title = {ECLAIR: enhanced clarification for interactive responses in an enterprise AI assistant},
year = {2025},
isbn = {978-1-57735-897-8},
publisher = {AAAI Press},
url = {https://doi.org/10.1609/aaai.v39i28.35363},
doi = {10.1609/aaai.v39i28.35363},
booktitle = {Proceedings of the Thirty-Ninth AAAI Conference on Artificial Intelligence and Thirty-Seventh Conference on Innovative Applications of Artificial Intelligence and Fifteenth Symposium on Educational Advances in Artificial Intelligence},
articleno = {3465},
numpages = {3},
series = {AAAI'25/IAAI'25/EAAI'25}
}

@misc{ambLLM,
      title={Ambiguity in LLMs is a concept missing problem}, 
      author={Zhibo Hu and Chen Wang and Yanfeng Shu and Hye-Young Paik and Liming Zhu},
      year={2025},
      eprint={2505.11679},
      archivePrefix={arXiv},
      primaryClass={cs.CL},
      url={https://arxiv.org/abs/2505.11679}, 
}

@article{huang2022language,
      title={Language Models as Zero-Shot Planners: Extracting Actionable Knowledge for Embodied Agents},
      author={Huang, Wenlong and Abbeel, Pieter and Pathak, Deepak and Mordatch, Igor},
      journal={arXiv preprint arXiv:2201.07207},
      year={2022}
}

@InProceedings{Johnson_2015_CVPR,
author = {Johnson, Justin and Krishna, Ranjay and Stark, Michael and Li, Li-Jia and Shamma, David and Bernstein, Michael and Fei-Fei, Li},
title = {Image Retrieval Using Scene Graphs},
booktitle = {Proceedings of the IEEE Conference on Computer Vision and Pattern Recognition (CVPR)},
month = {June},
year = {2015}
}

@ARTICLE{SG_image_caption,
  author={Yang, Xu and Zhang, Hanwang and Cai, Jianfei},
  journal={IEEE Transactions on Pattern Analysis and Machine Intelligence}, 
  title={Auto-Encoding and Distilling Scene Graphs for Image Captioning}, 
  year={2022},
  volume={44},
  number={5},
  pages={2313-2327},
  doi={10.1109/TPAMI.2020.3042192}}

@InProceedings{Nguyen_2021_ICCV,
    author    = {Nguyen, Kien and Tripathi, Subarna and Du, Bang and Guha, Tanaya and Nguyen, Truong Q.},
    title     = {In Defense of Scene Graphs for Image Captioning},
    booktitle = {Proceedings of the IEEE/CVF International Conference on Computer Vision (ICCV)},
    month     = {October},
    year      = {2021},
    pages     = {1407-1416}
}

@misc{sceneGraphVQA,
      title={Scene Graph Reasoning for Visual Question Answering}, 
      author={Marcel Hildebrandt and Hang Li and Rajat Koner and Volker Tresp and Stephan Günnemann},
      year={2020},
      eprint={2007.01072},
      archivePrefix={arXiv},
      primaryClass={cs.LG},
      url={https://arxiv.org/abs/2007.01072}, 
}

@INPROCEEDINGS{sceneGraphGen,
  author={Johnson, Justin and Gupta, Agrim and Fei-Fei, Li},
  booktitle={2018 IEEE/CVF Conference on Computer Vision and Pattern Recognition}, 
  title={Image Generation from Scene Graphs}, 
  year={2018},
  volume={},
  number={},
  pages={1219-1228},
  doi={10.1109/CVPR.2018.00133}}

@INPROCEEDINGS{TaskandMot1,
  author={Srivastava, Siddharth and Fang, Eugene and Riano, Lorenzo and Chitnis, Rohan and Russell, Stuart and Abbeel, Pieter},
  booktitle={2014 IEEE International Conference on Robotics and Automation (ICRA)}, 
  title={Combined task and motion planning through an extensible planner-independent interface layer}, 
  year={2014},
  volume={},
  number={},
  pages={639-646},
  doi={10.1109/ICRA.2014.6906922}}

@ARTICLE{Text2R,
  author={Yang, Zejun and Ning, Li and Wang, Haitao and Jiang, Tianyu and Zhang, Shaolin and Cui, Shaowei and Jiang, Hao and Li, Chunpeng and Wang, Shuo and Wang, Zhaoqi},
  journal={IEEE Robotics and Automation Letters}, 
  title={Text2Reaction : Enabling Reactive Task Planning Using Large Language Models}, 
  year={2024},
  volume={9},
  number={5},
  pages={4003-4010},
  doi={10.1109/LRA.2024.3371223}}

@misc{FeedbackGrounding,
      title={Grounding LLMs For Robot Task Planning Using Closed-loop State Feedback}, 
      author={Vineet Bhat and Ali Umut Kaypak and Prashanth Krishnamurthy and Ramesh Karri and Farshad Khorrami},
      year={2024},
      eprint={2402.08546},
      archivePrefix={arXiv},
      primaryClass={cs.RO},
      url={https://arxiv.org/abs/2402.08546}, 
}

@inproceedings{bird,
  title={Bird's-Eye-View Scene Graph for Vision-Language Navigation},
  author={Liu, Rui and Wang, Xiaohan and Wang, Wenguan and Yang, Yi},
  booktitle={ICCV},
  pages={10968--10980},
  year={2023}
}

@inproceedings{
        rana2023sayplan,
        title={SayPlan: Grounding Large Language Models using 3D Scene Graphs for Scalable Task Planning},
        author={Krishan Rana and Jesse Haviland and Sourav Garg and Jad Abou-Chakra and Ian Reid and Niko Suenderhauf},
        booktitle={7th Annual Conference on Robot Learning},
        year={2023},
        url={https://openreview.net/forum?id=wMpOMO0Ss7a}
      }

@ARTICLE{SGsurvey,
  author={Chang, Xiaojun and Ren, Pengzhen and Xu, Pengfei and Li, Zhihui and Chen, Xiaojiang and Hauptmann, Alex},
  journal={IEEE Transactions on Pattern Analysis and Machine Intelligence}, 
  title={A Comprehensive Survey of Scene Graphs: Generation and Application}, 
  year={2023},
  volume={45},
  number={1},
  pages={1-26},
  doi={10.1109/TPAMI.2021.3137605}}

@article{roboexp,
    title={RoboEXP: Action-Conditioned Scene Graph via Interactive Exploration for Robotic Manipulation},
    author={Jiang, Hanxiao and Huang, Binghao and Wu, Ruihai and Li, Zhuoran and Garg, Shubham and Nayyeri, Hooshang and Wang, Shenlong and Li, Yunzhu},
    journal={CoRL},
    year={2024}
}

@article{sgbot,

  title={SG-Bot: Object Rearrangement via Coarse-to-Fine Robotic Imagination on Scene Graphs},

  author={Zhai, Guangyao and Cai, Xiaoni and Huang, Dianye and Di, Yan and Manhardt, Fabian and Tombari, Federico and Navab, Nassir and Busam, Benjamin},

journal={arXiv preprint arXiv:2309.12188},

}

@misc{verigraph,
      title={VeriGraph: Scene Graphs for Execution Verifiable Robot Planning}, 
      author={Daniel Ekpo and Mara Levy and Saksham Suri and Chuong Huynh and Abhinav Shrivastava},
      year={2024},
      eprint={2411.10446},
      archivePrefix={arXiv},
      primaryClass={cs.RO},
      url={https://arxiv.org/abs/2411.10446}, 
}

@InProceedings{SGnoLLM,
  title = 	 {Self-Supervised Learning of Scene-Graph Representations for Robotic Sequential Manipulation Planning},
  author =       {Nguyen, Son and Oguz, Ozgur and Hartmann, Valentin and Toussaint, Marc},
  booktitle = 	 {Proceedings of the 2020 Conference on Robot Learning},
  pages = 	 {2104--2119},
  year = 	 {2021},
  editor = 	 {Kober, Jens and Ramos, Fabio and Tomlin, Claire},
  volume = 	 {155},
  series = 	 {Proceedings of Machine Learning Research},
  month = 	 {16--18 Nov},
  publisher =    {PMLR},
  url = 	 {https://proceedings.mlr.press/v155/nguyen21b.html}
}

@inproceedings{SGnoLLM2,
author = {Zhu, Yifeng and Tremblay, Jonathan and Birchfield, Stan and Zhu, Yuke},
title = {Hierarchical Planning for Long-Horizon Manipulation with Geometric and Symbolic Scene Graphs},
year = {2021},
publisher = {IEEE Press},
url = {https://doi.org/10.1109/ICRA48506.2021.9561548},
doi = {10.1109/ICRA48506.2021.9561548},
booktitle = {2021 IEEE International Conference on Robotics and Automation (ICRA)},
pages = {6541–6548},
numpages = {8},
location = {Xi'an, China}
}

@INPROCEEDINGS{SGnoLLM3, 
              AUTHOR    = {Abdelrhman Werby AND Chenguang Huang AND Martin Büchner AND Abhinav Valada AND Wolfram Burgard}, 
              TITLE     = {{Hierarchical Open-Vocabulary 3D Scene Graphs for Language-Grounded Robot Navigation}}, 
              BOOKTITLE = {Proceedings of Robotics: Science and Systems}, 
              YEAR      = {2024}, 
              ADDRESS   = {Delft, Netherlands}, 
              MONTH     = {July}, 
              DOI       = {10.15607/RSS.2024.XX.077} 
          }

@article{yao2022react,
  title={ReAct: Synergizing Reasoning and Acting in Language Models},
  author={Yao, Shunyu and Zhao, Jeffrey and Yu, Dian and Du, Nan and Shafran, Izhak and Narasimhan, Karthik and Cao, Yuan},
  journal={arXiv preprint arXiv:2210.03629},
  year={2022}
}

@article{yin2024sgnav, 
          title={SG-Nav: Online 3D Scene Graph Prompting for LLM-based Zero-shot Object Navigation}, 
          author={Hang Yin and Xiuwei Xu and Zhenyu Wu and Jie Zhou and Jiwen Lu},
          journal={arXiv preprint arXiv:2410.08189},
          year={2024}
    }

@ARTICLE{Lemma,
  author={Gong, Ran and Gao, Xiaofeng and Gao, Qiaozi and Shakiah, Suhaila and Thattai, Govind and Sukhatme, Gaurav S.},
  journal={IEEE Robotics and Automation Letters}, 
  title={LEMMA: Learning Language-Conditioned Multi-Robot Manipulation}, 
  year={2023},
  volume={8},
  number={10},
  pages={6835-6842},
  doi={10.1109/LRA.2023.3313058}}

@article{GDINO,
  title={Grounding dino: Marrying dino with grounded pre-training for open-set object detection},
  author={Liu, Shilong and Zeng, Zhaoyang and Ren, Tianhe and Li, Feng and Zhang, Hao and Yang, Jie and Li, Chunyuan and Yang, Jianwei and Su, Hang and Zhu, Jun and others},
  journal={arXiv preprint arXiv:2303.05499},
  year={2023}
}

@inproceedings{art,
    title = "The Art of {SOCRATIC} {QUESTIONING}: Recursive Thinking with Large Language Models",
    author = "Qi, Jingyuan  and
      Xu, Zhiyang  and
      Shen, Ying  and
      Liu, Minqian  and
      Jin, Di  and
      Wang, Qifan  and
      Huang, Lifu",
    editor = "Bouamor, Houda  and
      Pino, Juan  and
      Bali, Kalika",
    booktitle = "Proceedings of the 2023 Conference on Empirical Methods in Natural Language Processing",
    month = dec,
    year = "2023",
    address = "Singapore",
    publisher = "Association for Computational Linguistics",
    url = "https://aclanthology.org/2023.emnlp-main.255/",
    doi = "10.18653/v1/2023.emnlp-main.255",
    pages = "4177--4199"
}

@inproceedings{cot,
author = {Wei, Jason and Wang, Xuezhi and Schuurmans, Dale and Bosma, Maarten and Ichter, Brian and Xia, Fei and Chi, Ed H. and Le, Quoc V. and Zhou, Denny},
title = {Chain-of-thought prompting elicits reasoning in large language models},
year = {2022},
isbn = {9781713871088},
publisher = {Curran Associates Inc.},
address = {Red Hook, NY, USA},
booktitle = {Proceedings of the 36th International Conference on Neural Information Processing Systems},
articleno = {1800},
numpages = {14},
location = {New Orleans, LA, USA},
series = {NIPS '22}
}

@article{gemini,
  title={Gemini 2.5: Pushing the frontier with advanced reasoning, multimodality, long context, and next generation agentic capabilities},
  author={Comanici, Gheorghe and Bieber, Eric and Schaekermann, Mike and Pasupat, Ice and Sachdeva, Noveen and Dhillon, Inderjit and Blistein, Marcel and Ram, Ori and Zhang, Dan and Rosen, Evan and others},
  journal={arXiv preprint arXiv:2507.06261},
  year={2025}
}

@misc{qwen,
      title={Qwen3 Technical Report}, 
      author={An Yang and Anfeng Li and Baosong Yang and Beichen Zhang and Binyuan Hui and Bo Zheng and Bowen Yu and Chang Gao and Chengen Huang and Chenxu Lv and Chujie Zheng and Dayiheng Liu and Fan Zhou and Fei Huang and Feng Hu and Hao Ge and Haoran Wei and Huan Lin and Jialong Tang and Jian Yang and Jianhong Tu and Jianwei Zhang and Jianxin Yang and Jiaxi Yang and Jing Zhou and Jingren Zhou and Junyang Lin and Kai Dang and Keqin Bao and Kexin Yang and Le Yu and Lianghao Deng and Mei Li and Mingfeng Xue and Mingze Li and Pei Zhang and Peng Wang and Qin Zhu and Rui Men and Ruize Gao and Shixuan Liu and Shuang Luo and Tianhao Li and Tianyi Tang and Wenbiao Yin and Xingzhang Ren and Xinyu Wang and Xinyu Zhang and Xuancheng Ren and Yang Fan and Yang Su and Yichang Zhang and Yinger Zhang and Yu Wan and Yuqiong Liu and Zekun Wang and Zeyu Cui and Zhenru Zhang and Zhipeng Zhou and Zihan Qiu},
      year={2025},
      eprint={2505.09388},
      archivePrefix={arXiv},
      primaryClass={cs.CL},
      url={https://arxiv.org/abs/2505.09388}, 
}

\end{document}